\documentclass[lettersize,journal]{IEEEtran}
\usepackage{amsmath,amsfonts}
\usepackage{algorithm}
\usepackage{algorithmic}
\usepackage{array}
\usepackage{graphicx}
\usepackage[caption=false,font=normalsize,labelfont=sf,textfont=sf]{subfig}
\usepackage{textcomp}
\usepackage{stfloats}
\usepackage{url}
\usepackage{verbatim}
\usepackage{graphicx}
\usepackage{xcolor}
\usepackage{cite}
\usepackage{amssymb}
\usepackage{makecell}
\usepackage{booktabs} 
\usepackage[table]{xcolor} 
\usepackage{booktabs}

\usepackage{soul, color, xcolor} 
\sethlcolor{yellow} 

\usepackage{graphicx}
\usepackage{enumitem}
\usepackage{booktabs}
\usepackage{pifont}
\newcommand{\cmark}{\ding{51}}
\newcommand{\xmark}{\ding{55}}

\soulregister{\cite}7 
\soulregister{\citep}7 
\soulregister{\citet}7 
\soulregister{\ref}7 
\soulregister{\pageref}7 

\begin{document}

\title{OR-VSKC: Resolving Visual-Semantic Knowledge Conflicts in Operating Rooms with Synthetic Data-Guided Alignment}

\author{Weiyi~Zhao,
        Xiaoyu~Tan,
        Liang~Liu,
        Sijia~Li,
        Youwei~Song,
        and~Xihe~Qiu*
\thanks{W. Zhao, X. Tan, and L. Liu contributed equally to this work. }%
\thanks{Corresponding author: Xihe Qiu.(e-mail: qiuxihe1993@gmail.com)}
\thanks{W. Zhao, S. Li, Y. Song, and X. Qiu are with the Shanghai University of Engineering Science, Shanghai, China.}%
\thanks{X. Tan is with Tencent YouTu Lab, Shanghai, China.}%
\thanks{L. Liu is with the Clinical Research Unit, Zhongshan Hospital of Fudan University, Shanghai, China.}%
}

\markboth{Journal of \LaTeX\ Class Files,~Vol.~14, No.~8, August~2021}%
{Zhao \MakeLowercase{\textit{et al.}}: Visual-Semantic Knowledge Conflicts in Operating Room Scenes} 

\markboth{Journal of IEEE Transactions on Multimedia}%
{Shell \MakeLowercase{\textit{et al.}}: A Sample Article Using IEEEtran.cls for IEEE Journals}


\maketitle

\begin{abstract}
Automated identification of surgical safety risks is critical for improving patient outcomes; however, Multimodal Large Language Models (MLLMs) frequently suffer from Visual-Semantic Knowledge Conflicts (VS-KC), a phenomenon where models possess safety knowledge but fail to activate it during visual inspection. Investigating this alignment gap in operating rooms (ORs) is impeded by a critical bottleneck: the scarcity and privacy constraints of real-world OR data depicting safety violations. To address this, we introduce OR-VSKC, a benchmark for studying VS-KC and surgical risk perception in strictly regulated OR environments. Constructed via our Protocol-to-Pixel Generative Framework, OR-VSKC comprises 28,190 high-fidelity synthetic images grounded in authoritative safety standards, complemented by a 713-image expert-authored challenge subset validated by multiple experts. The full benchmark is built from authentic OR contexts drawn from the 4D-OR and CAMMA-MVOR datasets, where the 4D-OR-based portion serves as the primary benchmark core and the CAMMA-MVOR-based portion is reserved for external validation and cross-dataset generalization analysis. Evaluations of state-of-the-art MLLMs reveal substantial reliability gaps even in advanced generalist models. Furthermore, experiments show that fine-tuning on OR-VSKC effectively mitigates VS-KC and enables robust generalization to unseen camera viewpoints. We open-source the code and dataset to support reproducible research in safety-critical medical environments. The source code is available at \url{https://github.com/zgg2577/VS-KC}.
\end{abstract}

\section{Introduction}
\begin{figure}[htbp] 
    \centering
    \includegraphics[width=0.8\linewidth]{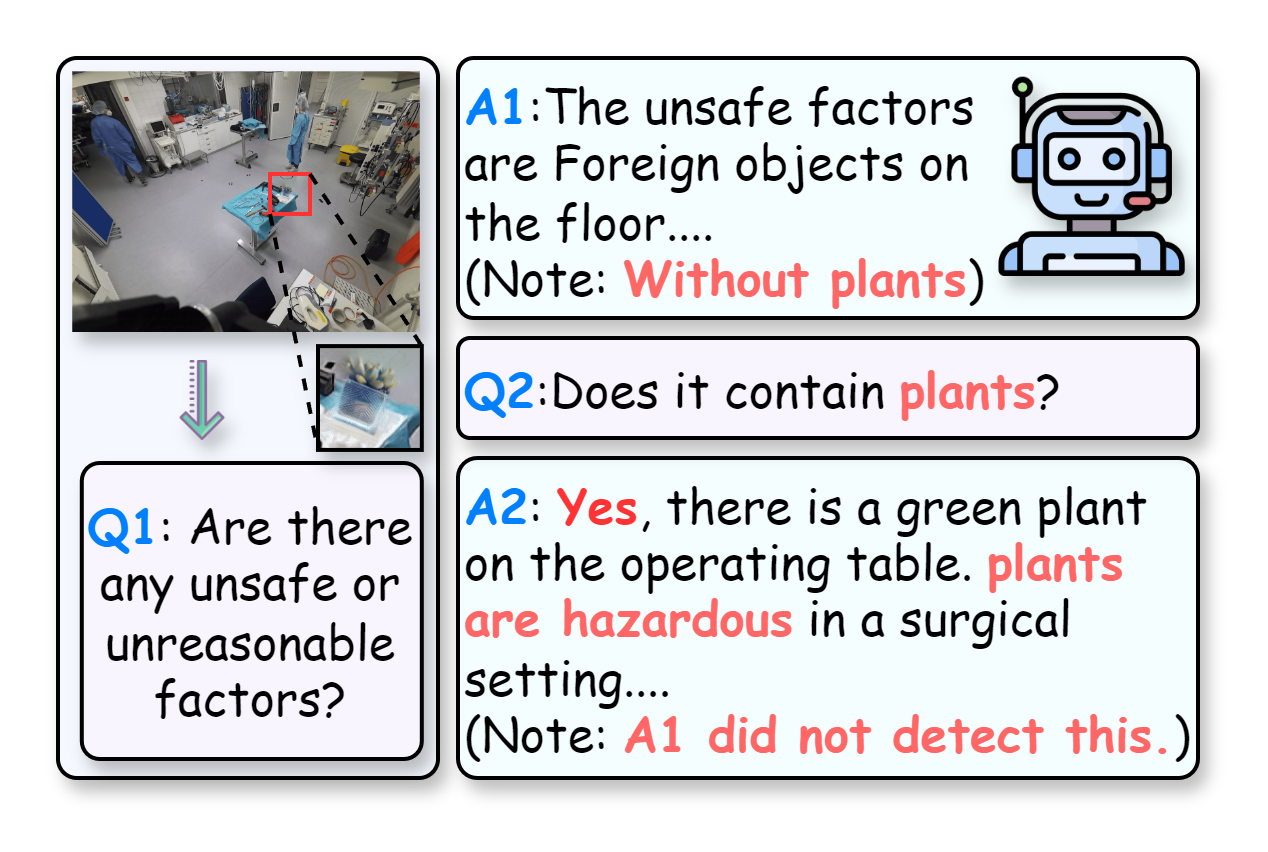} 
    \caption{Example of visual-semantic knowledge conflicts}
    \label{fig:vskc_example}
\end{figure}

\begin{figure*}[htbp]
    \centering
    \includegraphics[width=0.9\textwidth]{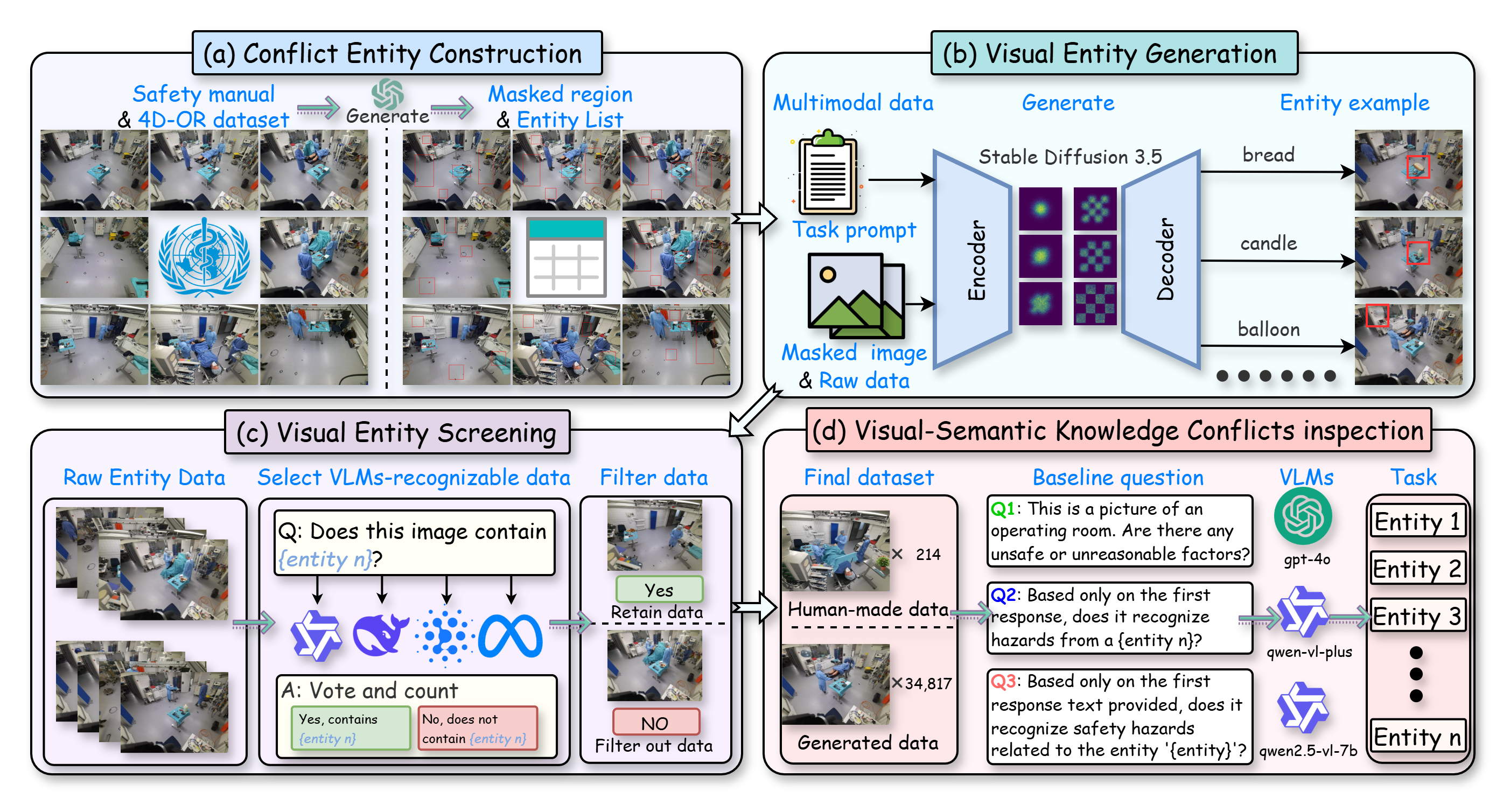}
    \caption{Overview of the OR-VSKC Dataset Generation and VS-KC Inspection Framework: 
     (a) Constructing specifications for visual conflict entities and their target locations within OR scenes, guided by safety protocols and operating room image analysis. 
    (b) Synthesizing OR scenes by embedding conflict entities according to specifications, using a diffusion model. 
    (c) Filtering generated images by ensuring consistent entity detectability via a heterogeneous multi-model consensus process. 
    (d) Evaluating MLLMs for VS-KC on the final curated OR-VSKC dataset via a structured, multi-stage query protocol.}
    \label{fig:1}
\end{figure*}

Operating room (OR) safety is critical to surgical quality and patient outcomes. Despite continuous improvements in surgical protocols and monitoring systems, adverse events still occur in clinical practice, calling for accurate and timely risk identification~\cite{armstrong2022effect}. Recent advances in artificial intelligence, especially Multimodal Large Language Models (MLLMs), offer a promising technical path for visual perception, contextual understanding, and safety assessment in complex surgical environments\cite{kung2023performance}. Building on the success of LLMs in medical tasks ranging from education to perioperative logistics \cite{ramamurthi2025applying,ke2025clinical}, MLLMs synergize visual perception with linguistic reasoning to enhance decision-making in complex environments \cite{zhang2024mm,cui2024survey}. In the high-stakes context of the OR, these models offer a promising avenue for automated situational awareness \cite{xiao2024comprehensive}, with recent works exploring language-guided robotic assistance \cite{moghani2024sufia}, image-grounded question answering for specific surgeries \cite{he2024pitvqa}, and multi-view generalization via models like ORacle \cite{ozsoy2024oracle}.

Despite these advancements, effective MLLM deployment remains hindered by critical reliability issues, particularly inconsistencies with clinical reality \cite{pal2024gemini,bai2024hallucination}. We identify a distinct challenge within this domain: \textbf{Visual-Semantic Knowledge Conflicts (VS-KC)}. While MLLMs often articulate correct safety rules in text-only scenarios, they frequently fail to autonomously detect overt violations when presented with corresponding visual input. As illustrated in Figure~\ref{fig:vskc_example}, a model might overlook a hazardous plant in an operating room image during a general assessment, yet correctly identify the risk when specifically prompted. We hypothesize that this reflects a ``lazy safety'' phenomenon: the model possesses the \textit{latent competence} to recognize the hazard but fails to activate this knowledge during \textit{active visual scanning}. This perception-reasoning misalignment creates a critical reliability gap in rule-intensive environments.

Systematically mitigating VS-KC necessitates a dedicated benchmark to quantify the alignment gap between visual perception and safety reasoning. The operating room serves as an optimal zero-tolerance environment for such investigation, where strict protocols render safety violations unambiguous and logically binary. However, establishing an OR-based benchmark faces a critical bottleneck: the extreme scarcity of real-world violation data, compounded by privacy regulations, precludes large-scale collection. To bridge this gap, we introduce a \textbf{Protocol-to-Pixel Generative Framework}. Theoretically grounded in the principle of \textbf{eliminating perceptual confounding variables}, this pipeline utilizes context-aware adversarial injection to translate abstract sterile protocols into high-fidelity visual semantic traps. Specifically, to strictly evaluate semantic reasoning rather than low-level perceptual robustness, our framework operationalizes this principle through three mechanisms: \textbf{Rule-Grounded Entity Mapping}, which employs adversarial probing to align synthetic violations with clinical ground truth; \textbf{Resolution-Aware Spatial Constraints}, which decouple semantic understanding from perceptual acuity by enforcing geometric visibility to rule out small object detection failures; and \textbf{Consensus-Driven Visual Verification}, which leverages a heterogeneous expert committee to filter generative artifacts and ensure unambiguous visual evidence. By strictly controlling these spatial and semantic attributes, our framework rigorously isolates and exposes the VS-KC phenomenon without relying on restricted clinical data.

In this paper, we present the following contributions:
\begin{itemize}
    \item We introduce and formalize the concept of Visual-Semantic Knowledge Conflicts within surgical operating room environments. We propose a systematic framework that maps abstract safety protocols to visual hazards, employing adversarial MLLM probing to identify and expose critical blind spots in model reasoning.
    
    \item We propose a dedicated benchmark designed to quantitatively assess Visual-Semantic Knowledge Conflicts. Addressing the scarcity of domain-specific resources, we construct and open-source OR-VSKC, a large-scale benchmark comprising \textbf{28,190 machine-screened synthetic samples} and a \textbf{713-image expert-authored challenge subset}. This benchmark incorporates a hierarchical three-tiered cognitive taxonomy (Common, Occasional, and Unreasonable) and is validated through a hybrid quality assurance strategy: heterogeneous ensemble screening for the synthetic data and multi-expert validation for the expert-authored challenge subset. The source code and dataset are open-sourced\footnote{\url{https://github.com/zgg2577/VS-KC}}.
    
    \item We evaluate state-of-the-art MLLMs, revealing significant reliability gaps in current generalist models. Furthermore, we demonstrate that fine-tuning on OR-VSKC effectively mitigates these conflicts by aligning visual perception with safety protocols. Crucially, our experiments validate that this alignment facilitates robust generalization to unseen camera angles, proving the dataset's efficacy in teaching view-invariant surgical safety principles rather than mere memorization.
\end{itemize}

\section{Related Work}

Knowledge conflicts in large language models (LLMs) have been widely studied as a major cause of inconsistent and unreliable model behavior when accessible knowledge contains contradictory information. Xu \emph{et al.}~\cite{xu2024knowledge} categorize knowledge conflicts into three types according to their sources: Context-Memory conflict, where external contextual information contradicts the model's internal parametric knowledge; Inter-Context conflict, where multiple contextual sources are mutually inconsistent; and Intra-Memory conflict, where contradictions exist within the model's own stored knowledge. Intra-Memory conflict can manifest as divergent outputs for semantically equivalent inputs and may arise from biased or heterogeneous training corpora, stochastic decoding, imperfect knowledge editing, unstable latent knowledge representations, and cross-lingual inconsistencies~\cite{wang2023causal,xu2022does,lee2022factuality,huang2025survey,yao2023editing,li2023unveiling,chuang2023dola,qi2023cross}. Existing mitigation strategies usually aim to improve consistency through consistency-aware fine-tuning, validator models, plug-in mechanisms, knowledge editing, or output ensembling~\cite{elazar2021measuring,mitchell2022enhancing}.

In this work, visual-semantic knowledge conflict is formulated as a multimodal form of Context-Memory conflict. Specifically, the observed surgical scene provides dynamic visual context, while the model's parametric knowledge contains general and domain-specific operating-room safety rules. A VS-KC occurs when visually recognizable entities appear plausible at the perception level but contradict protocol-grounded safety knowledge. Therefore, the challenge is not limited to object recognition or visual question answering; rather, the model must determine whether the recognized visual content is semantically compatible with clinical safety protocols. Although VS-KC is mainly triggered by visual context conflicting with parametric safety knowledge, Intra-Memory conflict remains relevant because unstable or inconsistent internal safety knowledge can weaken the model's ability to judge whether a scene violates protocol requirements. This makes VS-KC particularly important in zero-tolerance operating-room scenarios, where visually ordinary but protocol-violating foreign entities may indicate potential safety hazards.

Existing clinical and operating-room multimodal datasets provide important foundations for medical visual perception, surgical scene understanding, pose estimation, and visual question answering. Nevertheless, these benchmarks mainly evaluate whether models can recognize visual content, infer spatial or structural relations, or answer clinically relevant questions. They do not directly examine whether a model can detect a semantic conflict between visually plausible entities and protocol-grounded safety rules. This limitation is critical in operating-room scenarios, where a correct recognition result may still be insufficient if the model fails to identify the recognized entity as a safety violation. As summarized in Table~\ref{tab:dataset_comparison}, OR-VSKC addresses this gap by reframing evaluation from passive perception and VQA toward active protocol-grounded conflict detection under zero-tolerance surgical safety constraints.

\begin{table*}[t]
\centering
\fontsize{8pt}{9.6pt}\selectfont
\caption{COMPARISON OF OR-VSKC WITH REPRESENTATIVE CLINICAL MULTIMODAL DATASETS.}
\label{tab:dataset_comparison}
\resizebox{\textwidth}{!}{%
\begin{tabular}{l c l c c}
\toprule
\rowcolor{gray!20}
\textbf{Dataset} & 
\textbf{Scale} & 
\textbf{Task Focus} & 
\makecell{\textbf{Protocol-Grounded}\\\textbf{Conflict Entities}} & 
\makecell{\textbf{Direct VS-KC}\\\textbf{Evaluation}} \\
\midrule

\rowcolor{gray!6}
4D-OR \cite{ozsoy20224d}
& 6.7k fr.
& OR scene graphs, human/object poses, clinical roles
& \xmark
& \xmark \\

MVOR \cite{srivastav2018mvor}
& 732 mv-fr.
& Multi-view human detection and 2D/3D pose estimation in OR
& \xmark
& \xmark \\

\rowcolor{gray!6}
PitVQA \cite{he2024pitvqa}
& 25 vid.
& Surgical visual question answering in pituitary surgery
& \xmark
& \xmark \\

SLAKE \cite{liu2021slake}
& 642 img.
& Medical visual question answering with semantic labels
& \xmark
& \xmark \\

\rowcolor{gray!6}
VQA-RAD \cite{lau2018dataset}
& 315 img.
& Radiology visual question answering
& \xmark
& \xmark \\

\rowcolor{blue!10}
\textbf{OR-VSKC (ours)}
& 28.2k syn. + 713 chal.
& Protocol-grounded foreign-entity safety conflicts in OR
& \cmark
& \cmark \\

\bottomrule
\end{tabular}%
}
\end{table*}

\section{Methodology}\label{sec:Generation}

\subsection{Visual-Semantic Knowledge Conflicts Formulation}
\label{subsec:formulation}
VS-KC represent a fundamental misalignment between perceptual processing and semantic reasoning in multimodal large language models. We formalize this phenomenon within a set-theoretic framework. Let $\mathcal{M}$ denote a multimodal large language model, $\mathcal{V}$ be the visual input space, and $\mathcal{R}$ be a set of domain-specific safety rules. For a given scene $I \in \mathcal{V}$, let $\mathcal{O}(I)$ represent the ground-truth set of objects present in the scene. A safety violation occurs if there exists an entity $e \in \mathcal{O}(I)$ such that $e$ violates a rule $r \in \mathcal{R}$.

We define the operational capabilities of $\mathcal{M}$ through two functional projections:
\begin{itemize}
\item \textbf{Active Perception ($f_{\text{scan}}$):} The model's ability to autonomously identify hazards under a general safety prompt $q_{\text{scan}}$ (e.g., ``Are there unsafe factors?'').
\item \textbf{Latent Competence:} The model's underlying capability to recognize specific entities (via $f_{\text{comp}}$) and recall safety rules (via $f_{\text{sem}}$) when explicitly queried with focused prompts $q_{\text{verify}}$ and $q_{\text{rule}}$, respectively.
\end{itemize}

Formally, a Visual-Semantic Knowledge Conflict is identified when a hazard entity $e \in \mathcal{O}(I)$ exists, yet the model exhibits a divergence between its active perception and latent competence. This is characterized by the satisfaction of two conditions.
First, \textbf{active detection failure} occurs when:
\begin{equation}
e \notin f_{\text{scan}}(I, q_{\text{scan}}),
\end{equation}
indicating that the model fails to retrieve or report the hazard during a standard safety workflow.
Second, \textbf{competence retention} is demonstrated by:
\begin{equation}
\left( e \in f_{\text{comp}}(I, q_{\text{verify}}) \right) \land \left( f_{\text{sem}}(q_{\text{rule}}) \to \text{True} \right).
\end{equation}
Equation (2) distinguishes VS-KC from perceptual blindness or ignorance. It confirms that the model \textit{can} visually perceive the entity $e$ (when prompted specifically) and \textit{does} possess the semantic knowledge that $e$ is hazardous (when queried textually). The conflict, therefore, is not a lack of capability but a failure of cross-modal triggering: the visual presence of $e$ fails to activate the associated risk rule $r$ without explicit textual guidance.

\subsection{Conflict Entity Construction}
To systematically instantiate the conflicts formalized in Section~\ref{subsec:formulation}, we construct the OR-VSKC dataset via a mapping function $\Gamma: \mathcal{R} \rightarrow \mathcal{E} \times \mathcal{L}$, as illustrated in Figure~\ref{fig:1}(a). This function maps a semantic safety rule $r \in \mathcal{R}$ to a visual tuple $(e, l)$, where $e \in \mathcal{E}$ denotes a physical conflict entity, such as a non-sterile personal item, and $l \in \mathcal{L}$ represents its restricted location, such as the surgical field. The complete construction pipeline is detailed in Algorithm~\ref{alg:or_vskc_construction}.

The construction of the entity space $\mathcal{E}$ follows a selection protocol designed to maximize diagnostic value. The domain of rules $\mathcal{R}$ is derived from authoritative safety standards, including the WHO Best Practice Safety Protocols\cite{WHO_SurgicalSafety} and NatSSIPs\cite{CPOC_NatSSIPs}, ensuring that every entity represents a grounded safety threat. To bridge the gap between abstract protocols and specific visual challenges, We employed an adversarial MLLM probing strategy. Specifically, we queried multiple frontier models, including DeepSeek, Qwen, GPT, and Gemini, with prompts designed to surface entities that are visually salient and readily interpretable by humans, yet likely to induce visual-semantic blind spots in MLLMs. The core probing question asked the models to reason about which additional entities could be inserted into an operating room image such that a human observer would readily recognize them as hazardous or inappropriate, while a large multimodal model might still misjudge, overlook, or under-report them during general safety scanning. This multi-model probing stage was used only for candidate generation, rather than as a benchmark evaluation procedure.

The resulting candidates were not accepted automatically. Instead, they were further consolidated and finalized through discussion with the expert panel. A candidate entity was retained only if it satisfied three screening criteria: (i) protocol grounding, meaning the entity must correspond to a plausible violation under authoritative OR safety rules; (ii) visual discernibility, meaning the entity must remain clearly recognizable in realistic OR scenes; and (iii) blind-spot inducibility, meaning the entity should be likely to trigger a meaningful gap between passive recognition and active hazard reporting. Furthermore, to ensure that model failures stem from reasoning deficits rather than perceptual limitations, candidate entities were filtered based on visual discernibility and generative feasibility. This ensured that selected entities $e$ remain recognizable within complex surgical backgrounds and could be rendered with high fidelity. This framework ensures that every synthetic sample is rooted in a ground-truth violation $r$, thereby satisfying the precondition for competence retention (Equation~2) and establishing a robust basis for the dataset composition described in Section~\ref{sec:dataset_composition}.

\subsection{Conflict Entity Generation}
To operationalize the abstract conflict tuples $(e, l)$ derived via the mapping function $\Gamma$, we instantiate the visual scenes using a generative inpainting transformation. For the complete OR-VSKC release, we utilize authentic frames from the 4D-OR and CAMMA-MVOR datasets as the base canvas $I_{\text{base}}$, thereby enhancing environmental diversity across operating room layouts, viewpoints, and scene configurations. In the experimental protocol, however, the 4D-OR-based portion serves as the primary benchmark core, while the CAMMA-MVOR-based portion is reserved for external validation and cross-dataset generalization analysis. We define a synthesis function $\Phi$ that transforms $I_{\text{base}}$ into a synthetic image $I_{\text{syn}}$ containing the instantiated hazard:
\begin{equation}
I_{\text{syn}} = \Phi(I_{\text{base}}, M_l, P_e; \theta_{\text{SD}}),
\end{equation}
where $\theta_{\text{SD}}$ represents the frozen parameters of the Stable Diffusion 3.5 Medium model\cite{esser2024scaling,rombach2022high}. Here, $M_l$ denotes the spatial inpainting mask corresponding to the target location $l$, and $P_e$ represents the prompt conditioning derived from $e$, which comprises both positive entity descriptions and negative constraints to suppress generative artifacts.

The generation pipeline, depicted in Figure~\ref{fig:1}(b), employs a dual-control strategy to maximize visual coherence and recognizability.
First, regarding \textbf{spatial constraints}, we enforce a minimum resolution threshold where the area of the mask satisfies:
\begin{equation}
|M_l| > 0.05 \cdot |I_{\text{base}}|.
\end{equation}
This geometric constraint ensures that the generated entity occupies at least 5\% of the visual field. This threshold is critical to preclude detection failures attributed to low resolution or small object sizes, ensuring that any subsequent recognition failure is semantic rather than perceptual.

Second, for \textbf{visual fidelity}, the diffusion process employs a high denoising strength (0.90) and guidance scale (7.5) within the masked region. This configuration aims to reconstruct the entity $e$ while blending it into the lighting and texture distribution of the authentic OR environment $I_{\text{base}}$. However, due to the stochastic nature of diffusion models, this generative process does not guarantee perfect perceptual quality. The resulting raw candidate dataset, $\mathcal{D}_{\text{raw}} = \{I_{\text{syn}}^{(i)}\}_{i=1}^N$, inevitably contains instances of generative artifacts, ambiguous object rendering, or unnatural blending. These visual inconsistencies necessitate a robust downstream verification mechanism to ensure that the introduced conflicts are clearly discernible before being used for evaluation.

\subsection{Visual Entity Screening}
Given that the raw samples in $\mathcal{D}_{\text{raw}}$ may suffer from the perceptual inconsistencies noted above, direct inclusion would compromise the validity of the benchmark. To refine this dataset, we apply a rigorous filtering function $\Psi$. While standard methods often rely on a single verifier, such approaches are prone to hallucinating positive detections on poor-quality images. To address this, we introduce a heterogeneous ensemble protocol that demands cross-model consensus to confirm visual validity.

\paragraph{Heterogeneous Expert Selection}
We construct an expert committee $\mathcal{C} = \{M_1, ..., M_K\}$ comprising $K=4$ distinct models: three generalist VLMs (Qwen2.5-VL-7b\cite{Qwen2VL,qwen2.5-VL,bai2025qwen25vltechnicalreport}, GLM-4.6V-Flash, Llama-3.2-11B-Vision-Instruct) and one specialized DeepSeek-OCR\cite{wei2025deepseek} model. This heterogeneity is critical; by employing models with different architectures and training distributions, we mitigate the risk of architecture-specific blind spots or shared hallucinations regarding the generated entities.

\paragraph{Consensus-Driven Verification}
For each candidate image $I_{\text{syn}}$, we query each expert $m \in \mathcal{C}$ with a verification prompt $q_{\text{verify}}$: ``Does this image contain $e$?''. The filtering function $\Psi(I_{\text{syn}})$ acts as a gatekeeper:
\begin{equation}
\Psi(I_{\text{syn}}) =
\begin{cases}
1 & \text{if } \sum_{m \in \mathcal{C}} \mathbb{I}(m(I_{\text{syn}}, q_{\text{verify}}) = \text{Yes}) \ge \tau \\
0 & \text{otherwise},
\end{cases}
\end{equation}
where $\mathbb{I}(\cdot)$ is the indicator function and $\tau=3$ is the majority consensus threshold. We set $\tau=3$ (allowing for a single dissenting vote) rather than requiring unanimity ($K=4$) to accommodate minor stochastic variances in individual model predictions. This threshold serves to balance filtration rigor with dataset diversity, ensuring retained samples are visually salient without being overly penalized by isolated expert hallucinations. Only images where $\Psi(I_{\text{syn}})=1$ are retained, forming the final curated dataset $\mathcal{D}_{\text{curated}}$.

\begin{algorithm}[!t]
\small
\setlength{\textfloatsep}{4pt}
\setlength{\floatsep}{4pt}
\setlength{\intextsep}{4pt}
\setlength{\abovecaptionskip}{2pt}
\setlength{\belowcaptionskip}{2pt}
\caption{Construction Pipeline of the OR-VSKC Dataset}
\label{alg:or_vskc_construction}
\begin{algorithmic}[1]
\setlength{\itemsep}{0pt}
\setlength{\parskip}{0pt}
\REQUIRE Safety Rules $\mathcal{R}$, Base Frames $\mathcal{D}_{\text{base}}$, Inpainting Model $\Phi(\cdot; \theta_{\text{SD}})$, Expert Committee $\mathcal{C} = \{M_1, \dots, M_K\}$, Verification Prompt Template $q_{\text{verify}}(\cdot)$, Consensus Threshold $\tau$
\ENSURE Curated Synthetic Dataset $\mathcal{D}_{\text{final}}$

\STATE \textsc{Phase 1: Conflict Entity Construction}
\STATE $\mathcal{T} \gets \emptyset$
\FORALL{$r \in \mathcal{R}$}
    \STATE $(e, l) \gets \text{AdversarialProbe}(r)$ \hfill \textit{\{Identify blind spots via MLLM\}}
    \IF{$\text{IsDiscernible}(e) \land \text{IsFeasible}(e) \land \text{IsBlindSpotInducing}(e)$}
        \STATE $\mathcal{T} \gets \mathcal{T} \cup \{(e, l)\}$ \hfill \textit{\{Map via $\Gamma : \mathcal{R} \rightarrow \mathcal{E} \times \mathcal{L}$\}}
    \ENDIF
\ENDFOR

\STATE \textsc{Phase 2: Conflict Entity Generation}
\STATE $\mathcal{D}_{\text{raw}} \gets \emptyset$
\FORALL{$(e, l) \in \mathcal{T}$}
    \FORALL{$I_{\text{base}} \in \mathcal{D}_{\text{base}}$}
        \STATE $M_l \gets \text{GenerateMask}(I_{\text{base}}, l)$
        \IF{$|M_l| > 0.05 \cdot |I_{\text{base}}|$}
            \STATE $P_e \gets \text{ConstructPrompt}(e)$
            \STATE $I_{\text{syn}} \gets \Phi(I_{\text{base}}, M_l, P_e; \theta_{\text{SD}})$ \hfill \textit{\{Inpainting synthesis (Eq.~(3))\}}
            \STATE $\mathcal{D}_{\text{raw}} \gets \mathcal{D}_{\text{raw}} \cup \{(I_{\text{syn}}, e, l)\}$ \hfill \textit{\{Spatial constraint in Eq.~(4)\}}
        \ENDIF
    \ENDFOR
\ENDFOR

\STATE \textsc{Phase 3: Visual Entity Screening}
\STATE $\mathcal{D}_{\text{final}} \gets \emptyset$
\FORALL{$(I_{\text{syn}}, e, l) \in \mathcal{D}_{\text{raw}}$}
    \STATE $\text{Votes} \gets \emptyset$
    \FOR{$k = 1$ to $K$}
        \STATE $v_k \gets M_k(I_{\text{syn}}, q_{\text{verify}}(e))$ \hfill \textit{\{Ask: ``Does this image contain $e$?''\}}
        \STATE $\text{Votes} \gets \text{Votes} \cup \{v_k\}$
    \ENDFOR
    \IF{$\sum_{k=1}^{K} \mathbf{1}(v_k = \texttt{Yes}) \ge \tau$}
        \STATE $\mathcal{D}_{\text{final}} \gets \mathcal{D}_{\text{final}} \cup \{(I_{\text{syn}}, e, l)\}$ \hfill \textit{\{Apply filtering $\Psi$ as in Eq.~(5)\}}
    \ENDIF
\ENDFOR

\STATE \textbf{return} $\mathcal{D}_{\text{final}}$
\end{algorithmic}
\vspace{-1mm}
\end{algorithm}

\subsection{Evaluation Metric}
\label{sec:eval_metric}
To quantify the prevalence of VS-KC, we define a structured evaluation protocol that measures the discrepancy between active risk perception and passive entity recognition. For a target model $\mathcal{M}$, we probe two distinct functional states.

\paragraph{Active Hazard Perception ($R_{\text{active}}$)}
This metric captures the model's autonomous ability to identify risks without guidance. It is defined as the textual response sequence generated by the model when presented with the open-ended safety scanning prompt $q_{\text{scan}}$: ``This is a picture of an operating room. Are there any unsafe or unreasonable factors? Answer briefly.'' Let $R_{\text{active}}$ denote this raw string output corresponding to $f_{\text{scan}}(I, q_{\text{scan}})$.

\paragraph{Passive Entity Recognition ($R_{\text{passive}}$)}
This metric serves as a control variable to verify the model’s visual competence. It is defined as the textual response sequence elicited by the targeted existence prompt $q_{\text{verify}}$: ``Does this image contain $e$?'' Let $R_{\text{passive}}$ denote this output corresponding to $f_{\text{comp}}(I, q_{\text{verify}})$.

To rigorously evaluate these linguistic outputs, we employ a self-reflective judge function, denoted as $\mathcal{J}(R, e)$. Crucially, we design this verification phase as a unimodal textual assessment. The judge model receives only the generated text response $R$ and the query regarding entity $e$, strictly excluding the original visual input $I$. This isolation ensures that the judgment is performed purely on the semantic level, preventing visual features from confounding the logic verification.

Rather than relying on an external oracle, we employ the target model $\mathcal{M}$ to evaluate the response $R$, where $R \in \{R_{\text{active}}, R_{\text{passive}}\}$. No additional training or optimization is applied to the judge; instead, the target MLLM operates in an inference-only, self-reflective setting. By formulating the evaluation as a textual entailment task, which is less complex than the original multimodal reasoning problem, we leverage the linguistic capabilities of the model to ascertain whether the generated response semantically entails hazard recognition. Crucially, because visual input is omitted at this stage, the judge assesses only the semantic content of the previously generated textual response, rather than perceiving the scene anew. This design avoids introducing an auxiliary model and mitigates the risk of the verification step being compromised by novel multimodal visual hallucinations. Although this self-reflective evaluation does not entirely eliminate circularity, it serves as a pragmatic and internally consistent verification strategy within a simplified unimodal context. The judge employs a strict binary classification prompt: ``Based only on the provided response, does it recognize hazards from $e$? Answer strictly with Yes or No.''

We define the VS-KC resolution score ($S_{\text{res}}$) for a given sample as the successful alignment of active visual detection and semantic reporting:
\begin{equation}
S_{\text{res}} = \mathcal{J}(R_{\text{active}}, e).
\end{equation}
Ideally, a safety-critical model should achieve $S_{\text{res}}=1$. The VS-KC phenomenon is strictly observed when the model fails the active check yet passes the passive check. Formally, a conflict is registered if and only if:
\begin{equation}
\text{Conflict} \iff \mathcal{J}(R_{\text{active}}, e) = 0 \quad \land \quad \mathcal{J}(R_{\text{passive}}, e) = 1.
\end{equation}
This metric structure effectively decouples active safety reasoning from passive visual recognition. By isolating cases where visual grounding succeeds but risk assessment fails, we specifically quantify the ``lazy safety'' behavior characteristic of VS-KC. Accordingly, the \textit{Accuracy} reported in all experimental tables is computed as the mean VS-KC resolution score over the evaluated samples:
\begin{equation}
\text{Accuracy} = \frac{1}{N}\sum_{i=1}^{N} S_{\text{res}}^{(i)} \times 100\%
= \frac{1}{N}\sum_{i=1}^{N} \mathcal{J}(R_{\text{active}}^{(i)}, e_i) \times 100\%.
\end{equation}
Category-wise accuracies are computed within each semantic category, while the overall accuracy is computed over all samples in the corresponding evaluation subset.

\section{Experimental Results}
\label{sec:results}

\begin{figure}[t]
\centering
\includegraphics[width=0.8\linewidth]{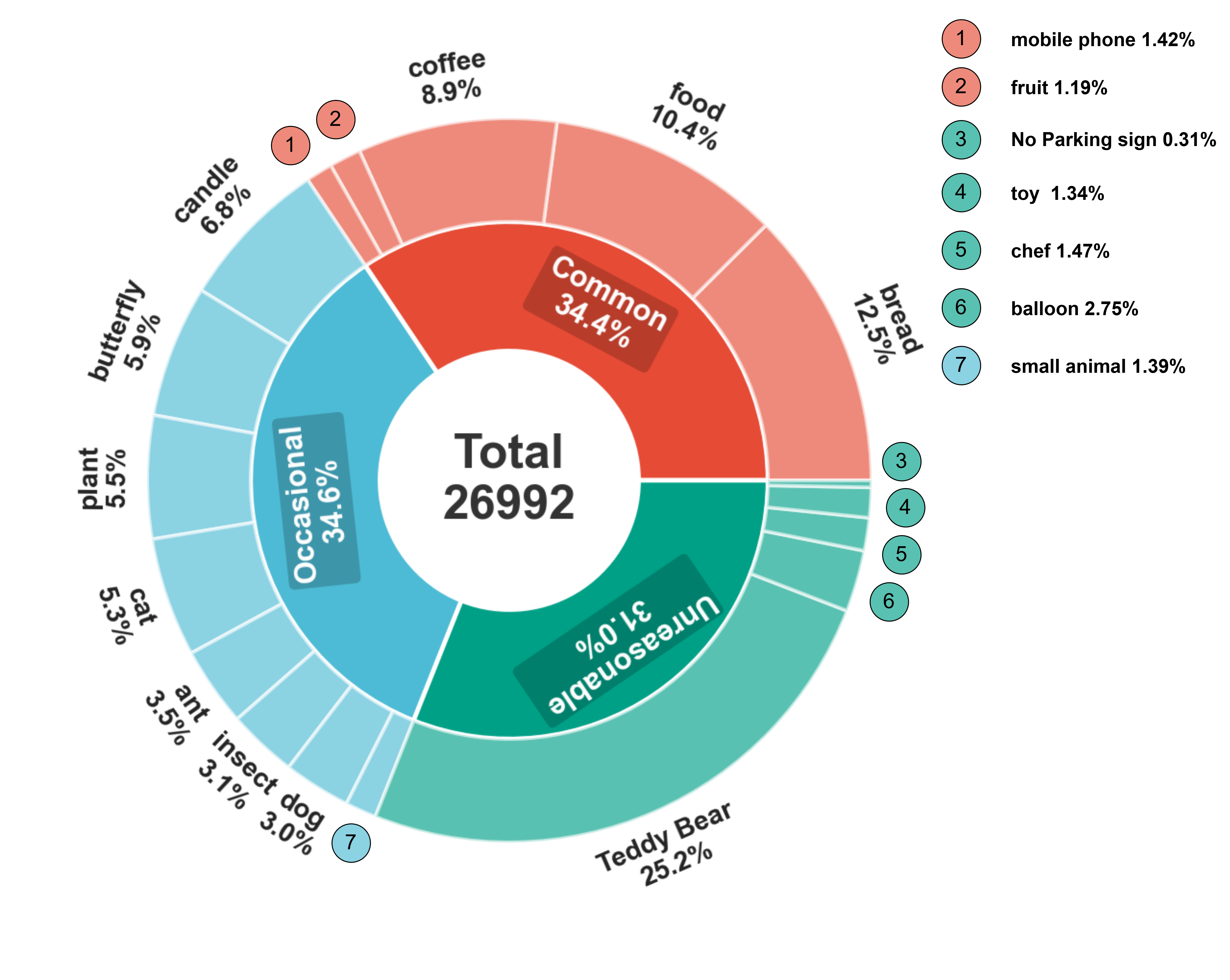}
\caption{Hierarchical distribution of conflict entities in the synthetic OR-VSKC subset constructed from 4D-OR scenes (N = 26,992). The inner ring illustrates the distribution across three semantic plausibility levels: Common (34.4\%), Occasional (34.6\%), and Unreasonable (31.0\%). The outer ring delineates the specific entity types within each respective level. This organizational structure provides comprehensive coverage of OR safety violations, ranging from high-frequency protocol breaches to semantically unreasonable conflicts.}
\label{fig:data_distribution}
\end{figure}

\subsection{Dataset Composition}
\label{sec:dataset_composition}

We introduce OR-VSKC, a benchmark designed to facilitate research on VS-KC in operating room scene recognition. The benchmark comprises two complementary components: a large-scale, machine-screened synthetic subset and an expert-authored challenge subset for rigorous evaluation. These components are structured according to their source domains. Specifically, we partition the dataset into a 4D-OR-based core subset, utilized for primary quantitative analyses, and a CAMMA-MVOR-based subset, reserved for external validation and cross-dataset generalization. To comprehensively capture temporal and contextual variations in the operating room, the selected source frames encompass multiple phases of the surgical workflow, including preoperative, intraoperative, and postoperative periods.

\subsubsection{The Synthetic OR-VSKC Dataset}

The full synthetic release consists of $N=28{,}190$ curated images. These were synthesized by applying our diffusion-based inpainting method (Section~\ref{sec:Generation}) to authentic operating-room frames drawn from the \textbf{4D-OR} and \textbf{CAMMA-MVOR} datasets, followed by the ensemble verification protocol detailed in Section~III-D. Within this full release, the \textbf{4D-OR}-based core synthetic subset contains $N=26,992$ images and is used for the primary benchmark statistics and the main quantitative analyses reported in this paper, while the \textbf{CAMMA-MVOR}-based synthetic subset contains $N=1,198$ images and is reserved for external validation and cross-dataset generalization analysis. This organization preserves a unified public benchmark while explicitly separating in-distribution evaluation from out-of-distribution validation. 

To evaluate the nuances of VS-KC, the \textbf{4D-OR}-based core synthetic subset encompasses 18 distinct entity types distributed across three semantic plausibility levels. As illustrated in Figure~\ref{fig:data_distribution}, this core synthetic subset maintains a balanced distribution across these categories. This three-tiered hierarchy is designed to progressively challenge MLLM capabilities, moving from specific procedural knowledge to foundational common sense.

Critically, in the high-stakes context of an operating room, the distinction between ``object detection'' and ``risk perception'' is functionally unified by sterile field protocols. Unlike general open-world scenarios where an object's risk depends on complex interactions, in a surgical zone, the mere presence of any non-sterile foreign entity constitutes an immediate safety violation. Therefore, our dataset operationalizes risk not through ambiguous activities, but through the verified physical presence of prohibited items, ensuring that successful detection is clinically equivalent to risk identification.

The conflict entities are categorized as follows:

a) Common Mistakes (High-Frequency Violations)
Accounting for 34.4\% of the dataset, this category includes items strictly prohibited but plausibly introduced due to negligence, such as personal electronics (\textit{mobile phone}: 1.42\%) and consumables (\textit{bread}, \textit{coffee}, \textit{fruit}, \textit{food}). This tier evaluates the model's ability to apply context-dependent rules and overcome ``normalcy bias,'' wherein common objects are perceived as benign regardless of setting. Failure here indicates a gap in specific domain knowledge rather than general recognition capabilities.

b) Occasional Mistakes (Environmental Breaches)
Representing 34.6\% of the data, this category features rarer but conceivable lapses in environmental control, including biological contaminants (\textit{ant}, \textit{cat}, \textit{insect}, \textit{plant}: 5.5\%) and fire hazards (\textit{candles}). Unlike common errors, these dynamic entities challenge the model to perform causal reasoning regarding environmental integrity. Success in this tier requires identifying ``out-of-context'' edge cases that threaten the sterile field, testing the ability to handle low-probability, high-consequence safety events.

c) Unreasonable Conflicts (Semantic Absurdities)
Constitutes 31.0\% of the dataset, featuring entities that defy the logical context of a surgical suite, such as irrelevant personnel (\textit{chef}), festive items (\textit{balloon}), and incongruous objects (\textit{Teddy Bear}: 25.2\%, \textit{No Parking sign}). This tier serves as a stress test for foundational common sense and hallucination resistance. Detecting these conflicts requires the model to understand the basic ``grammar'' of a scene, specifically its purpose and actor roles, and recognize the deep semantic incompatibility between these entities and the medical environment.

\subsubsection{The Expert-Authored and Multi-Expert-Validated Challenge Subset}
To complement the synthetic release, a 713-image expert-validated challenge subset was constructed for rigorous evaluation. For experimental clarity, this challenge subset is explicitly partitioned according to its source domains: a 4D-OR-based core subset comprising 509 samples utilized for primary benchmark analyses, and a CAMMA-MVOR-based subset of 204 samples reserved for external validation and cross-dataset generalization. The complete subset was independently created by a six-member interdisciplinary panel, comprising four clinical experts and two AI experts, using authentic operating room (OR) scene contexts. The lead clinical expert, an attending physician at the Clinical Research Institute of Zhongshan Hospital, Fudan University, possesses doctoral training and extensive experience in clinical oncology. This expert guided the assessment of clinical plausibility, patient-safety relevance, and semantic validity for the violation scenarios. The remaining clinical experts evaluated the consistency of the constructed scenarios with realistic clinical workflows and safety expectations of the operating room. The two AI experts, with backgrounds in multimodal learning and vision-language model evaluation, assessed visual construction quality, technical feasibility, and suitability for MLLM evaluation.

Candidate scenarios were initially defined according to safety rules of the operating room and aligned with the Common, Occasional, and Unreasonable taxonomy. The images were subsequently constructed through a hybrid workflow that combined manual compositing and expert-guided AI tools to insert conflict entities into authentic scene backgrounds, preserving spatial logic and visual clarity. Prior to independent review, the panel utilized a shared set of predefined criteria and taxonomy definitions to align the interpretation of clinical plausibility, visual unambiguity, violation semantics, and artifact exclusion. Consistency was promoted through these shared criteria, independent expert voting, and a majority-consensus retention rule.

Each candidate image was independently reviewed according to four criteria: clinical plausibility of the scene, visual clarity and unambiguity of the target entity, semantic clarity of the safety violation, and absence of confounding artifacts. A candidate sample was retained only upon satisfying these criteria and receiving approval from a minimum of four experts. The retained samples in the finalized challenge subset achieved a mean expert approval rate of 97.5\%. Based on the voting outcomes of the six raters for the binary retention decision, the inter-rater agreement reached a Fleiss kappa of $\kappa = 0.9636$, indicating near-perfect agreement. Because the scenario category (Common, Occasional, or Unreasonable) was determined by the predefined taxonomy during scenario definition rather than through open-ended post hoc labeling, inter-rater agreement was computed exclusively for the binary retention decision. This metric directly measures whether the experts consistently judged a candidate sample as clinically plausible, visually unambiguous, semantically valid, and free of artifacts.

By mirroring the categorical structure of the synthetic set across these partitions, this expert-validated challenge subset serves as a rigorous benchmark for evaluating clinical relevance and semantic robustness.

\begin{table}[!t]
\centering
\caption{MODEL PERFORMANCE AND ABLATION STUDY ON THE 4D-OR-BASED EXPERT-AUTHORED CHALLENGE SUBSET}
\label{tab:annotated_performance}
\small
\setlength{\tabcolsep}{5pt}
\renewcommand{\arraystretch}{1.2}
\resizebox{\columnwidth}{!}{%
\begin{tabular}{lcccc}
\toprule
\rowcolor{gray!20}
\textbf{Method} &
\makecell{\textbf{Common}\\($N$=150)} &
\makecell{\textbf{Occasional}\\($N$=229)} &
\makecell{\textbf{Unreasonable}\\($N$=130)} &
\makecell{\textbf{Overall}\\($N$=509)} \\
\midrule
GLM-4.6V-Flash~\cite{vteam2025glm45vglm41vthinkingversatilemultimodal} & 34.67\% & 31.44\% & 34.62\% & 33.20\% \\
Llama-3.2-11B-Vision-Instruct~\cite{grattafiori2024llama} & 63.33\% & 75.98\% & 63.85\% & 69.15\% \\
Qwen2.5-VL-7B-Instruct~\cite{bai2025qwen25vltechnicalreport} & 73.31\% & 57.73\% & 69.64\% & 65.36\% \\
BaichuanMed-OCR-7B~\cite{lijun2025BaichuanMed-OCR-7B} & 58.60\% & 67.21\% & 52.66\% & 60.96\% \\
Qwen-vl-plus~\cite{bai2025qwen25vltechnicalreport} & 57.35\% & 75.41\% & 52.46\% & 64.23\% \\
GPT-4o~\cite{hurst2024gpt} & 75.29\% & 78.40\% & 62.30\% & 73.37\% \\
Gemini-3-Flash~\cite{team2023gemini} & 82.00\% & 80.79\% & 73.08\% & 79.18\% \\
GPT-5.2~\cite{hurst2024gpt} & 90.67\% & 87.77\% & 81.54\% & 87.03\% \\
\midrule

Ours (FT, Qwen-Screen) & 91.63\% & \textbf{95.76\%} & 99.00\% & 95.37\% \\
\rowcolor{blue!10}
\textbf{Ours (FT, Ensemble)} & \textbf{95.86\%} & 95.61\% & \textbf{100.00\%} & \textbf{96.80\%} \\
\bottomrule
\end{tabular}%
}
\end{table}

\begin{table}[!t]
\centering
\caption{MODEL PERFORMANCE AND ABLATION STUDY BY CATEGORY ON THE SYNTHETIC OR-VSKC SUBSET CONSTRUCTED FROM 4D-OR}
\label{tab:synthetic_performance}
\fontsize{8pt}{9.6pt}\selectfont
\setlength{\tabcolsep}{5pt}
\renewcommand{\arraystretch}{1.2}
\resizebox{\columnwidth}{!}{%
\begin{tabular}{lcccc}
\toprule
\rowcolor{gray!20}
\textbf{Method} &
\textbf{Common} &
\textbf{Occasional} &
\textbf{Unreasonable} &
\textbf{Overall} \\
\midrule

GLM-4.6V-Flash & 34.26\% & 25.12\% & 37.25\% & 31.60\% \\
Llama-3.2-11B-Vision-Instruct & 62.19\% & 64.59\% & 64.03\% & 63.68\% \\
Qwen2.5-VL-7B-Instruct & 72.07\% & 56.79\% & 65.09\% & 64.01\% \\
BaichuanMed-OCR-7B & 69.83\% & 40.46\% & 56.75\% & 55.69\% \\
Qwen-vl-plus & 63.16\% & 65.48\% & 63.94\% & 64.40\% \\
GPT-4o & 71.75\% & 71.71\% & 60.14\% & 68.51\% \\

\midrule
Ours (FT, Qwen-Screen) & 92.14\% & \textbf{96.11\%} & 98.67\% & 95.54\% \\
\rowcolor{blue!10}
\textbf{Ours (FT, Ensemble)} & \textbf{99.27\%} & 95.14\% & \textbf{99.64\%} & \textbf{97.56\%} \\

\bottomrule
\end{tabular}%
}
\end{table}

\begin{table}[htbp]
\centering
\caption{Key Experimental Parameter Configurations}
\label{tab:experimental_param}
\scriptsize
\setlength{\tabcolsep}{5pt}
\renewcommand{\arraystretch}{1.05}
\begin{tabular}{
>{\centering\arraybackslash}m{0.22\columnwidth}
>{\raggedright\arraybackslash}m{0.68\columnwidth}
}
\toprule
\rowcolor{gray!25}
\multicolumn{1}{>{\centering\arraybackslash}m{0.22\columnwidth}}{\textbf{Phase}} &
\multicolumn{1}{>{\centering\arraybackslash}m{0.68\columnwidth}}{\textbf{Setting}} \\
\midrule

Visual Entity Screening
& MAX\_NEW\_TOKENS = 32; TEMPERATURE = 0.01 \\[2pt]

\rowcolor{gray!10}
LoRA Fine-tuning
& task\_type = ``CAUSAL\_LM''; inference\_mode = False; r = 64; 
LoRA\_alpha = 16; LoRA\_dropout = 0.05; learning\_rate = 1e-4; 
batch\_size = 4 \\[2pt]

\bottomrule
\end{tabular}
\end{table}

\subsection{Extended Experimental Setup}


Experiments were conducted on a server equipped with an 80GB GPU. Key parameter configurations are detailed in Table~\ref{tab:experimental_param}. The source code and dataset partitions have been open-sourced to support reproducibility.The primary synthetic dataset was divided into training and testing sets, with the latter comprising 9,397 samples. Additionally, the \textbf{713-image expert-authored challenge subset} was excluded from training and reserved as a high-fidelity benchmark for evaluation.

To assess generalization beyond the standard distribution, we analyzed a specific test subset of 1,735 images featuring \textit{unseen camera angles}. Unlike standard test samples that share training perspectives, this subset includes viewpoints withheld during the training phase. This setup evaluates whether the model acquires robust, view-invariant safety principles rather than relying on viewpoint-specific visual correlations.

\subsection{Experimental Setup and Baseline Analysis}

To assess contemporary MLLMs in detecting safety-critical conflicts, we benchmarked open-weight models (GLM-4.6V-Flash, Qwen2.5-VL-7B, Llama-3.2-11B-Vision-Instruct) and proprietary systems (Qwen-VL-Plus, GPT-4o, Gemini-3-Flash, GPT-5.2). Using our three-part taxonomy (Common, Occasional, Unreasonable), the evaluation tests contextual understanding across datasets (Tables~\ref{tab:annotated_performance} and~\ref{tab:synthetic_performance}). Crucially, the ``Accuracy'' reported in these analyses and tables directly corresponds to the mean VS-KC resolution score ($S_{res}$) defined in Section III-E.

We first established a comprehensive baseline on the expert-authored challenge subset to rigorously probe performance bounds regarding clinical reality. To this end, we deployed two additional frontier models, Gemini-3-Flash and GPT-5.2, exclusively on this verified ``gold standard'' subset to avoid prohibitive computational overhead. This high-fidelity benchmark revealed a significant shift in performance hierarchy. Gemini-3-Flash demonstrated superior efficiency, surpassing GPT-4o with higher accuracy in identifying Common Mistakes (82.00\% vs. 75.29\%). Furthermore, GPT-5.2 set a new baseline ceiling for generalist models, achieving $\sim$87\% average accuracy and narrowing the gap with domain-specific solutions. Among standard baselines, Llama-3.2-11B-Vision-Instruct emerged as a strong contender, outperforming Qwen-vl-plus with 75.98\% in Occasional Mistakes, suggesting specific strengths in handling realistic anomalies.

Following this high-fidelity evaluation, we assessed the standard baselines on the large-scale \textbf{synthetic dataset}. Here, GPT-4o demonstrated the most robust general capabilities among the standard models, achieving the highest weighted average accuracy of 68.51\% and excelling particularly in Common and Occasional Mistakes. Qwen-vl-plus followed closely, showing competitive performance (64.40\%), whereas the lightweight GLM-4.6V-Flash struggled significantly (31.60\%), highlighting the inherent difficulty of this task for smaller, efficiency-oriented models lacking domain adaptation.

Despite the strong performance of frontier models on the expert-authored baseline, a critical reliability gap remains. While GPT-5.2 dominated general benchmarks, it notably struggled with Unreasonable Conflicts (81.54\%), suggesting that even the most advanced generalist models harbor ``common sense hallucinations'' when facing logical absurdities in specialized contexts. In stark contrast, our fine-tuned model (Ours (FT)) achieved near-perfect performance, with averages of 96.80\% on expert-authored data and 97.56\% on synthetic data. Specifically, our model achieved 100.00\% on Unreasonable conflicts compared to GPT-5.2's 81.54\%, conclusively validating that general model scaling alone cannot replace the necessity of domain-specific training for zero-tolerance safety tasks.

\begin{table}[!t]
\centering
\caption{MODEL PERFORMANCE AND ABLATION STUDY ON GENERALIZATION TO UNSEEN CAMERA ANGLES}
\label{tab:ablation_unseen_angles}
\fontsize{8pt}{9.6pt}\selectfont
\setlength{\tabcolsep}{5pt}
\renewcommand{\arraystretch}{1.2}
\resizebox{\columnwidth}{!}{%
\begin{tabular}{lccc}
\toprule
\rowcolor{gray!20}
\textbf{Method} &
\textbf{Common} &
\textbf{Occasional} &
\textbf{Unreasonable} \\
\midrule

GLM-4.6V-Flash & 34.09\% & 25.87\% & 33.48\% \\
Qwen2.5-vl-7b (Base) & 60.39\% & 59.33\% & 52.69\% \\
Llama-3.2-11B-Vision-Instruct & 67.98\% & 80.51\% & 67.95\% \\
BaichuanMed-OCR-7B & 77.48\% & 78.69\% & 60.97\% \\
GPT-4o & 71.00\% & 69.28\% & 54.17\% \\

\midrule
Ours (FT, Qwen-Screen) & 87.19\% & 98.67\% & 99.14\% \\

\rowcolor{blue!10}
\textbf{Ours (FT, Ensemble)} & \textbf{96.08\%} & \textbf{99.19\%} & \textbf{99.65\%} \\

\bottomrule
\end{tabular}%
}
\end{table}

\subsection{Fine-tuning and Generalization Analysis}

The results presented in Table~\ref{tab:synthetic_performance} and Table~\ref{tab:annotated_performance} demonstrate a substantial performance improvement following fine-tuning. Our model (Ours (FT)) achieved average accuracies exceeding 97\% on the synthetic set and 95\% on the expert-authored validation set.

We interpret these results as the effective activation of the model's latent knowledge. Given that the conflict entities introduced in our dataset (e.g., mobile phones, teddy bears) possess high visual saliency and are extensively represented in general pre-training corpora, the model inherently possesses the perceptual capacity to detect them. The performance deficit of baseline models (e.g., Qwen2.5-VL-7B-Instruct at 64.01\%) was therefore not attributable to perceptual limitations, but rather to a disconnect between visual detection and risk assessment (the VS-KC phenomenon).

Fine-tuning effectively bridged this alignment gap by explicitly conditioning the model to map these foreign objects to safety protocols. Consequently, the model's performance naturally converged to its underlying visual recognition ceiling. Furthermore, the robust consistency observed on the expert-authored challenge subset (Table~\ref{tab:annotated_performance}) confirms that this alignment generalizes to real-world textures and is not merely an artifact of synthetic data.

\subsection{Ablation Study: Robustness to Unseen Angles}
\begin{figure}
    \centering
    \includegraphics[width=0.9\linewidth]{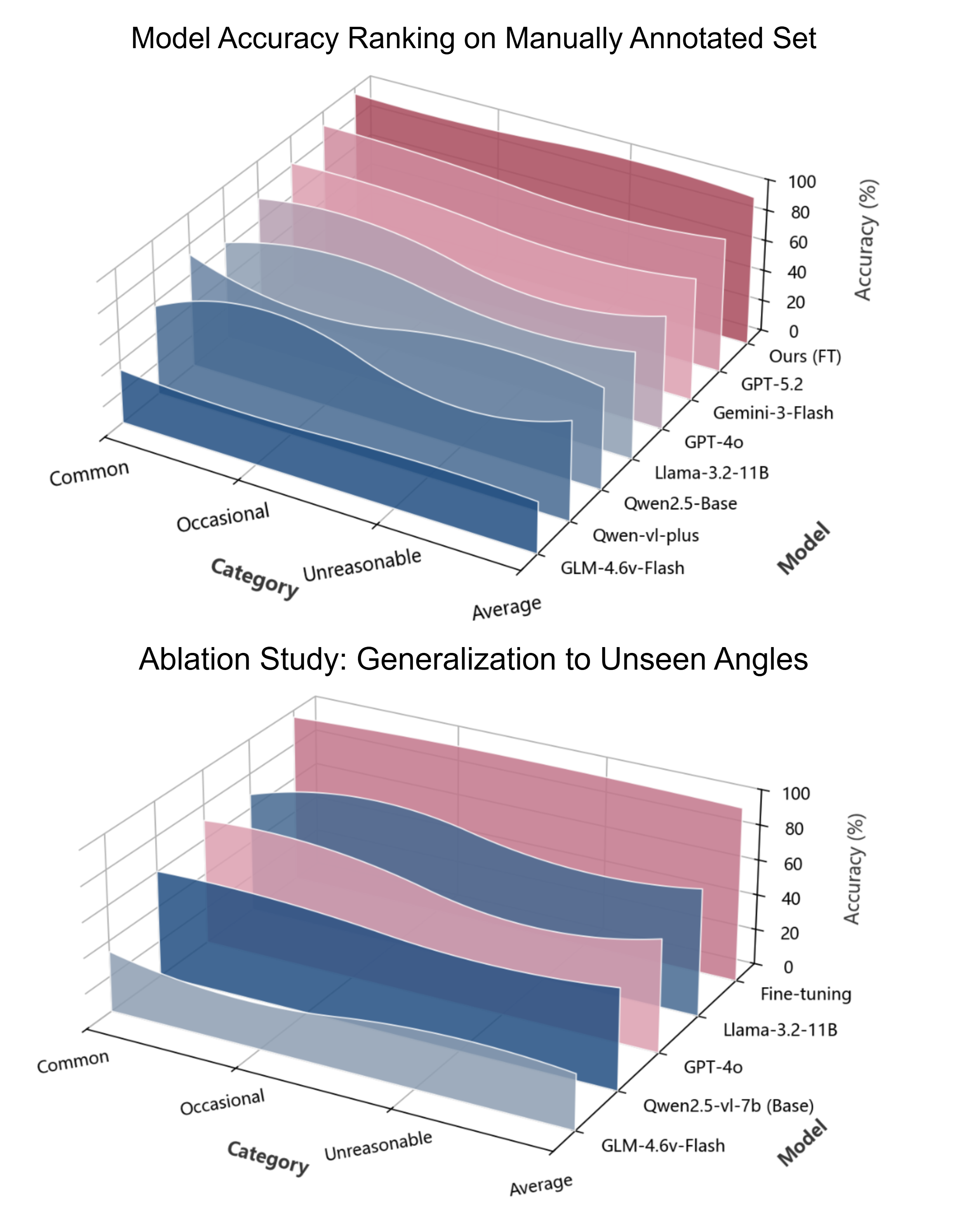}
    \caption{ Performance analysis of vision-language models on conflict recognition tasks}
    \label{fig:performance_analysis}
\end{figure}

To assess the generalization capabilities of our fine-tuned model, particularly its robustness to variations in visual perspective, we conducted a targeted robustness evaluation using images captured from camera angles explicitly excluded from the training data. We compared our fine-tuned model against five off-the-shelf baselines, including general-purpose MLLMs and one medical-domain multimodal baseline.(Table~\ref{tab:ablation_unseen_angles}). The lower panel of Fig.~\ref{fig:performance_analysis} provides a visual summary of this unseen-angle generalization comparison across different conflict categories.

The results demonstrate the substantial impact of our fine-tuning methodology. While the generalist baseline models, lacking domain-specific adaptation, struggled with the domain shift, exemplified by Qwen2.5-VL-7B-Instruct dropping to 56.94\% average accuracy, the Llama-3.2-11B-Vision-Instruct model exhibited notable resilience, particularly in the Occasional Mistakes category (80.51\%), where it outperformed GPT-4o (69.28\%).


Our fine-tuned model consistently outperformed all zero-shot baselines across all categories, with accuracy ranging from 96.08\% (Common) to 99.65\% (Unreasonable). These results suggest that the model learns viewpoint-invariant safety cues rather than memorizing training views. Its $>96\%$ accuracy on unseen camera angles further indicates strong generalization, which is important for safety assessment in dynamic clinical environments.

\subsection{Ablation Study: Effectiveness of Heterogeneous Ensemble Screening}

\begin{figure}[!t]
    \centering
    \includegraphics[width=0.9\linewidth]{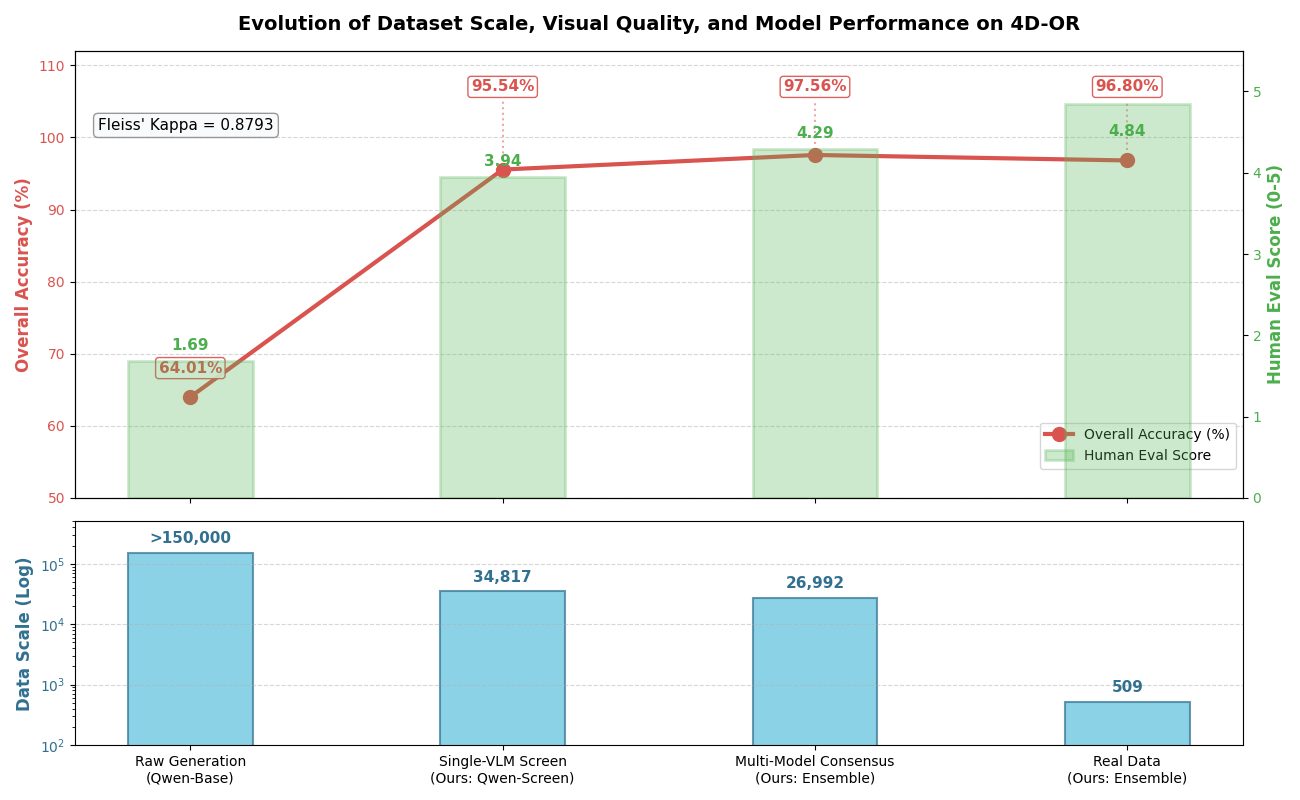}
    \caption{Impact of data curation stages on dataset scale, visual fidelity, and model alignment. This figure illustrates the inverse relationship between dataset volume (blue bars, logarithmic scale) and the corresponding improvements in visual plausibility (green bars) and safety perception accuracy (red line).}
    \label{fig:pipeline_analysis}
\end{figure}

To evaluate the effectiveness of the consensus-driven visual verification module in Section III-D, we compare two fine-tuning variants that differ only in the screening strategy used to construct the training data. \textbf{Ours (FT, Ensemble)} is trained on samples vetted by the heterogeneous multi-model committee, whereas \textbf{Ours (FT, Qwen-Screen)} is trained on samples filtered only by Qwen2.5-VL-7B-Instruct. This ablation isolates the effect of data curation quality from the model architecture and fine-tuning procedure.

The results show that heterogeneous ensemble screening provides more robust training data than single-model screening. Although \textbf{Ours (FT, Qwen-Screen)} already improves substantially over zero-shot baselines, it consistently underperforms the ensemble-screened variant under distribution shifts. On the CAMMA-MVOR expert-authored subset (Table~\ref{tab:camma_mvor_human_refined}), the overall accuracy decreases from 97.23\% to 90.19\% when replacing ensemble screening with Qwen-only screening, mainly due to the drop in \textit{Common} conflicts from 97.22\% to 81.99\%. A similar degradation is observed on the CAMMA-MVOR-based synthetic subset (Table~\ref{tab:camma_mvor_ai_refined}), where the overall accuracy decreases from 98.21\% to 86.96\%. These results suggest that single-model screening can introduce model-specific visual priors and retain ambiguous samples, while heterogeneous screening better suppresses such biases.

The category-level results further support this conclusion. Single-VLM screening remains relatively stable on \textit{Unreasonable} conflicts, which often contain visually and semantically obvious absurdities, but suffers larger drops on \textit{Common} conflicts, where high-frequency foreign objects are more susceptible to subtle generative artifacts and screening bias. Although the Qwen-screened variant slightly outperforms the ensemble variant in the \textit{Occasional} category on Table III (96.11\% vs. 95.14\%), this minor fluctuation is outweighed by its weaker robustness on unseen-angle and CAMMA-MVOR evaluations.

Fig.~\ref{fig:pipeline_analysis} further summarizes the curation process. The initial unconstrained generation stage produces over 150,000 raw synthetic images, which are reduced to 34,817 images by single-model screening. Heterogeneous ensemble screening further removes ambiguous samples, yielding a cleaner 4D-OR-based synthetic core subset of $N=26,992$ images. Human assessment confirms this quality improvement: visual plausibility increases from 1.69 for raw generations to 3.94 after single-model screening and 4.29 after heterogeneous ensemble curation, approaching the empirical reference level of authentic clinical images (4.84), with high inter-rater agreement (Fleiss $\kappa=0.8793$).

Overall, downstream performance correlates more strongly with data purity than with dataset scale. Without domain-specific alignment, \textbf{Qwen2.5-VL-7B-Instruct} achieves only 64.01\% accuracy. Fine-tuning on the Qwen-screened subset improves performance to 95.54\%, while fine-tuning on the ensemble-curated subset further increases accuracy to 97.56\% and maintains 96.80\% on the 509-image 4D-OR expert-authored challenge subset. These results demonstrate that heterogeneous ensemble screening serves as an effective regularizer, converting noisy synthetic generations into high-fidelity safety-alignment data and improving generalization across distinct OR data sources.

\begin{table}[!t]
\centering
\caption{MODEL PERFORMANCE ON THE CAMMA-MVOR-BASED SYNTHETIC SUBSET}
\label{tab:camma_mvor_ai_refined}
\small
\setlength{\tabcolsep}{5pt}
\renewcommand{\arraystretch}{1.2}
\resizebox{\columnwidth}{!}{%
\begin{tabular}{lcccc}
\toprule
\rowcolor{gray!20}
\textbf{Method} &
\makecell{\textbf{Common}\\($N$=599)} &
\makecell{\textbf{Occasional}\\($N$=475)} &
\makecell{\textbf{Unreasonable}\\($N$=124)} &
\textbf{Overall} \\
\midrule
BaichuanMed-OCR-7B & 76.51\% & 59.14\% & 7.41\% & 49.59\% \\
Llama-3.2-11B-Vision-Instruct & 45.40\% & 65.75\% & 22.04\% & 44.56\% \\
Qwen2.5-VL-7B-Instruct & 59.53\% & 64.92\% & 8.37\% & 42.52\% \\
GLM-4.6V-Flash & 93.33\% & 86.84\% & 69.72\% & 88.31\% \\
\midrule
\textbf{Ours (FT, Qwen-Screen)} & 82.90\% & 89.43\% & 97.14\% & 86.96\% \\
\rowcolor{blue!10}
\textbf{Ours (FT, Ensemble)} & \textbf{98.40\%} & \textbf{98.41\%} & \textbf{97.49\%} & \textbf{98.21\%} \\
\bottomrule
\end{tabular}%
}
\end{table}

\begin{table}[!t]
\centering
\caption{MODEL PERFORMANCE AND ABLATION STUDY ON THE EXPERT-AUTHORED CAMMA-MVOR CHALLENGE SUBSET}
\label{tab:camma_mvor_human_refined}
\small
\setlength{\tabcolsep}{5pt}
\renewcommand{\arraystretch}{1.2}
\resizebox{\columnwidth}{!}{%
\begin{tabular}{lcccc}
\toprule
\rowcolor{gray!20}
\textbf{Method} &
\makecell{\textbf{Common}\\($N$=57)} &
\makecell{\textbf{Occasional}\\($N$=84)} &
\makecell{\textbf{Unreasonable}\\($N$=63)} &
\makecell{\textbf{Overall}\\($N$=204)} \\
\midrule
BaichuanMed-OCR-7B & 68.08\% & 83.23\% & 9.65\% & 64.31\% \\
Llama-3.2-11B-Vision-Instruct & 3.80\% & 87.62\% & 23.75\% & 34.55\% \\
Qwen2.5-VL-7B-Instruct & 4.94\% & 87.37\% & 6.04\% & 30.92\% \\
GLM-4.6V-Flash & 95.53\% & 85.75\% & 59.35\% & 80.33\% \\
GPT-5.4-nano & 37.42\% & 34.33\% & 22.69\% & 31.60\% \\
GPT-5.4 & 98.00\% & 85.40\% & 91.10\% & 90.68\% \\
Gemini-3-flash & 90.92\% & 69.07\% & 79.05\% & 78.26\% \\
Gemini-3.1-flash & \textbf{98.75\%} & 86.41\% & 87.29\% & 90.13\% \\
\midrule

\textbf{Ours (FT, Qwen-Screen)} & 81.99\% & 90.55\% & 97.14\% & 90.19\% \\
\rowcolor{blue!10}
\textbf{Ours (FT, Ensemble)} & 97.22\% & \textbf{96.29\%} & \textbf{98.97\%} & \textbf{97.23\%} \\
\bottomrule
\end{tabular}%
}
\end{table}

\subsection{External Validation on the CAMMA-MVOR Subset}

To further strengthen the baseline comparison, we conduct external validation on the held-out \textbf{CAMMA-MVOR}-based expert-authored challenge subset and report the results in Table~\ref{tab:camma_mvor_human_refined}. This setting jointly probes cross-dataset transfer, medical-domain specialization, and the robustness of both open-source and proprietary MLLMs under a real OR source distinct from the primary 4D-OR-based benchmark core. All proprietary models were evaluated through their official APIs using greedy decoding, in order to reduce sampling variance and ensure deterministic comparison.

\textbf{BaichuanMed-OCR-7B} achieves a stronger overall score than several open-source general-purpose baselines (\textbf{64.31\%} vs. 34.55\% for Llama-3.2-11B-Vision-Instruct and 30.92\% for Qwen2.5-VL-7B-Instruct), showing that medical-domain multimodal pretraining benefits clinically relevant visual reasoning. This advantage is mainly reflected in the \textbf{Common} and \textbf{Occasional} categories, where BaichuanMed-OCR-7B attains 68.08\% and 83.23\%, respectively. However, its collapse on \textbf{Unreasonable} conflicts (\textbf{9.65\%}) indicates that generic medical specialization alone cannot reliably resolve VS-KC when protocol-grounded rejection of semantically incompatible entities is required.

The proprietary models further reveal the difficulty of OR-specific VS-KC reasoning. Although stronger closed-source models generally outperform open-source general-purpose baselines, their performance remains category-dependent. For example, \textbf{GPT-5.4} achieves a strong overall accuracy of \textbf{90.68\%}, with particularly high performance on \textbf{Common} and \textbf{Unreasonable} conflicts, while \textbf{Gemini-3.1-flash} reaches a comparable overall score of \textbf{90.13\%}. Nevertheless, these models still fall behind our fine-tuned model, especially in the \textbf{Occasional} and \textbf{Unreasonable} categories. This suggests that model scale and proprietary instruction tuning improve general visual-semantic reasoning, but do not fully eliminate the need for task-specific protocol grounding in safety-critical OR scenarios.

We also note the unusually strong CAMMA-MVOR performance of \textbf{GLM-4.6V-Flash}. This is likely a source-domain effect rather than a general safety-reasoning advantage: CAMMA-MVOR scenes may highlight inserted foreign entities, causing this lightweight model to rely on direct object-level cues under the open-ended safety-scanning prompt. Its weaker performance on \textbf{Unreasonable} conflicts and poor results on the 4D-OR core and unseen-angle evaluations further suggest that this advantage should not be viewed as robust protocol-grounded reasoning.

In contrast, our fine-tuned model remains uniformly strong across all categories, reaching \textbf{97.23\%} overall, with 97.22\%, 96.29\%, and 98.97\% on Common, Occasional, and Unreasonable conflicts, respectively. These results show that neither medical-domain pretraining, proprietary model scaling, nor incidental source-domain compatibility is sufficient for stable OR-specific safety violation detection. The gain of our model instead reflects the benefit of OR-VSKC and protocol-grounded alignment.

\subsection{Interpretability Analysis via Occlusion Sensitivity}

\begin{figure*}[!t]
    \centering
    \includegraphics[width=0.8\linewidth]{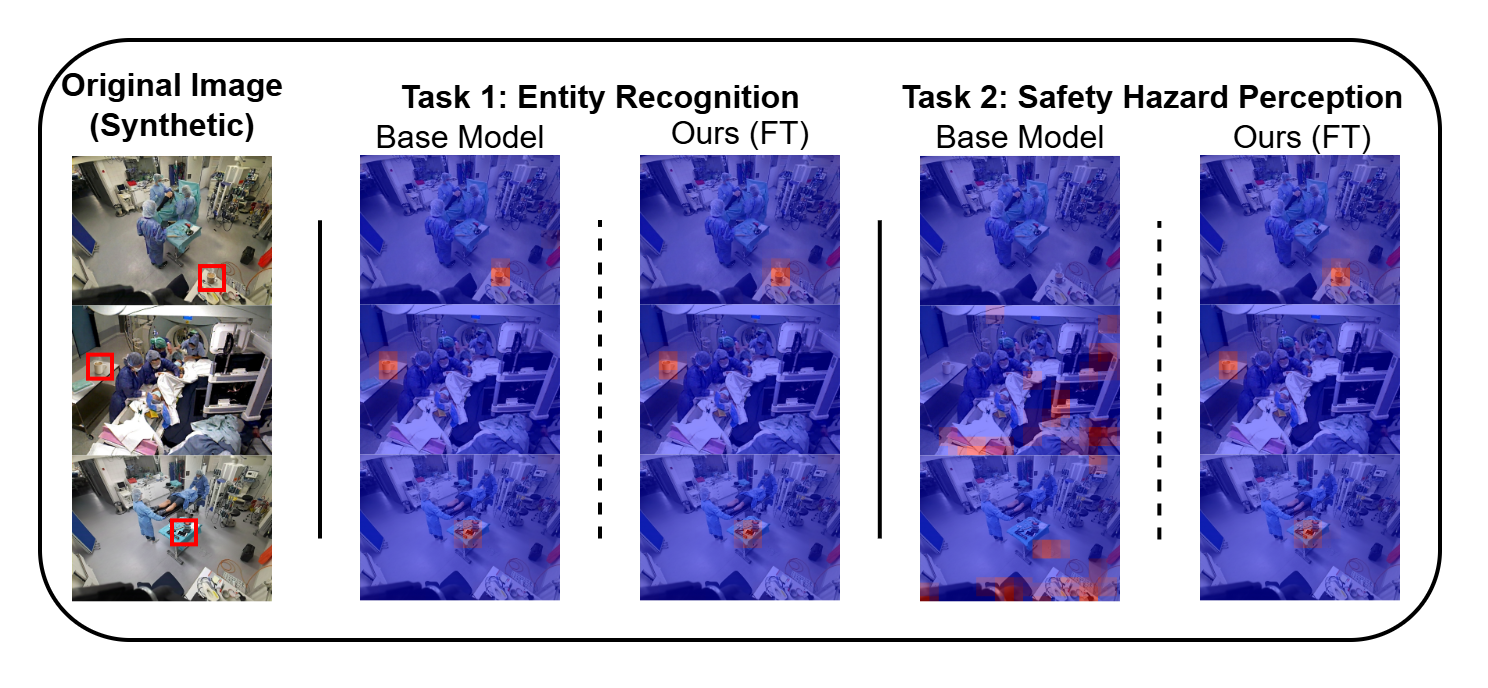} 
    \caption{\textbf{Interpreting Visual-Semantic Knowledge Conflicts via Occlusion Sensitivity.} Column 1 displays the original synthetic images with ground-truth bounds (red boxes). Row 1 and Row 2 show "coffee" entities across different OR scenes, while Row 3 shows a "mobile phone". \textbf{Task 1 (Cols 2-3):} Both models accurately localize the target when explicitly prompted for the specific entity. \textbf{Task 2 (Cols 4-5):} Under a safety scanning prompt, the Base Model (Col 4) exhibits severe attention scattering, failing to ground safety reasoning visually. In contrast, Ours (Col 5) precisely localizes the hazards, confirming successful alignment with surgical safety protocols.}
    \label{fig:occlusion_sensitivity}
\end{figure*}

To visually interpret the resolution of VS-KC, we conduct an occlusion sensitivity analysis (Fig. \ref{fig:occlusion_sensitivity}). This technique systematically obscures regions of the input image and measures the corresponding drop in the confidence of the model for a target prompt, effectively visualizing the semantic grounding of the decision-making process of the model.

Fig. \ref{fig:occlusion_sensitivity} presents a side-by-side comparison across two distinct tasks, featuring a coffee cup (Rows 1-2) and a mobile phone (Row 3) as representative conflict entities:

\textbf{Task 1: Entity Recognition}
Prompt: ``Does this image contain [Entity]?'': When explicitly queried about the presence of the specific entity, both the base model (Column 2) and our fine-tuned model (Column 3) generate tight, localized heatmaps over the ground-truth bounding boxes. This confirms that the baseline model possesses the necessary perceptual capacity to detect the foreign objects.
    
\textbf{Task 2: Active Hazard Perception}
Prompt $q_{\text{scan}}$: When presented with the open-ended safety scanning prompt defined in Section \ref{sec:eval_metric}, a significant difference is observed. The base model (Column 4) exhibits broad attention scattering, with its focus dispersed across irrelevant background regions such as floors and surgical drapes. This diffuse activation pattern illustrates the VS-KC phenomenon: despite perceiving the object, the model fails to associate it with the concept of a safety hazard. In contrast, our fine-tuned model (Column 5) maintains precise, pinpoint localization on the violating entities, consistent with its focus in Task 1.


This evidence shows that the failure of generalist models in OR safety tasks mainly arises from weak safety reasoning rather than poor object detection. Our protocol-grounded alignment strategy reduces this gap by linking active safety perception to clinical evidence of protocol violations.

\section{Discussion}
\label{sec:discussion}

The substantial performance improvement observed in our experiments indicates that the ``lazy safety'' phenomenon in MLLMs is primarily an alignment deficit rather than a perceptual inability. The strong detection performance achieved after fine-tuning, reaching above 97\% on the synthetic subset, suggests that fine-tuning effectively serves as a cognitive bridge, activating the pre-existing latent knowledge of the model. We ascribe these high saturation scores to the visual saliency of the selected conflict entities, such as mobile phones and teddy bears, which are extensively represented in pre-training data. This confirms that synthetic data is a highly sample-efficient strategy for grounding abstract safety rules into active visual triggers, transforming latent recognition capability into active risk perception for salient objects.

Despite these advancements, we acknowledge specific limitations regarding scope, generalization, and evaluation. First, the current definition of VS-KC focuses on the spatial existence of distinct foreign objects; more subtle hazards, such as temporal procedural errors or microscopic contamination, represent a more challenging and clinically complex frontier. Second, although the proposed method demonstrates robustness to unseen camera angles and a held-out external validation subset, the synthetic generation still relies on specific environmental priors, necessitating future cross-center validation to rule out domain-specific overfitting. The GLM-4.6V-Flash results on CAMMA-MVOR further suggest that lightweight MLLMs may exhibit source-domain-specific advantages when local visual saliency is amplified, but such advantages do not necessarily translate into robust semantic safety reasoning. Third, although the unimodal text-based self-evaluation minimizes circularity, deployment in high-stakes clinical settings requires broader heterogeneous expert verification to ensure reliability across diverse surgical scenarios.

We further note a practical constraint regarding clinical validation: unsafe intraoperative video frames are exceedingly scarce in real practice because operating room safety violations are strictly prohibited. In addition, collecting, curating, and releasing such data is constrained by substantial ethical, privacy, and patient safety considerations. As a result, prospective validation based on naturally occurring videos of actual violations is difficult to conduct in a rigorous and compliant manner. Under these constraints, the current validation strategy, which combines an expert-authored and multi-expert-validated challenge subset with held-out cross-dataset external validation, constitutes a clinically proximate and practically feasible evaluation setting. Nevertheless, prospective studies with surgical professionals on real intraoperative video streams would further strengthen the clinical translation of this work, and therefore remain an important direction for future research.

\section{Conclusion} 

This paper addresses the challenge of VS-KC in MLLMs within safety-critical operating room environments. To quantify this phenomenon, we introduce the \textbf{OR-VSKC benchmark}, constructed via our \textbf{Protocol-to-Pixel Generative Framework}. This benchmark integrates \textbf{28,190} high-fidelity synthetic images featuring rule-violating entities categorized by a three-tiered cognitive taxonomy, complemented by a \textbf{713-image expert-authored and multi-expert-validated challenge subset} for high-fidelity evaluation. Evaluations reveal critical reliability gaps in prominent generalist MLLMs. Fine-tuning on OR-VSKC aligns model perception with safety knowledge, improving conflict detection and generalization to unseen viewpoints. The dataset is publicly available. Although this work focuses on surgery, future studies will extend VS-KC to broader rule-based settings and develop stronger mitigation strategies for reliable MLLM deployment in safety-critical domains.


\bibliographystyle{IEEEtran}
\bibliography{ref}

\end{document}